\pdfoutput=1

\documentclass[11pt]{article}

\usepackage[final]{naacl2021}

\usepackage{times}
\usepackage{latexsym}

\usepackage[T1]{fontenc}

\usepackage[utf8]{inputenc}
\usepackage{microtype}
\usepackage{times}
\usepackage{latexsym}
\usepackage{graphicx}
\usepackage{subfiles}
\usepackage{float}
\usepackage{colortbl} 
\usepackage{booktabs}
\usepackage{multicol}
\usepackage{multirow}
\usepackage{listings}
\usepackage{dsfont}
\usepackage{mdframed}
\usepackage{amssymb}
\usepackage{enumitem}
\usepackage{amsmath}
\usepackage{soul}
\usepackage{mathptmx}
\usepackage{paracol}
\usepackage{xparse} 
\usepackage{fancyhdr}
\usepackage{subcaption}
\usepackage{todonotes}
\usepackage{diagbox}
\usepackage{comment}
\usepackage{microtype}
\usepackage{tikz,colortbl}

\usetikzlibrary{calc}
\usepackage{zref-savepos}

\newcounter{NoTableEntry}
\renewcommand*{\theNoTableEntry}{NTE-\the\value{NoTableEntry}}

\newcommand*{\notableentry}{%
  \multicolumn{1}{@{}c@{}|}{%
    \stepcounter{NoTableEntry}%
    \vadjust pre{\zsavepos{\theNoTableEntry t}}
    \vadjust{\zsavepos{\theNoTableEntry b}}
    \zsavepos{\theNoTableEntry l}
    \hspace{0pt plus 1filll}%
    \zsavepos{\theNoTableEntry r}
    \tikz[overlay]{%
      \draw[black]
        let
          \n{llx}={\zposx{\theNoTableEntry l}sp-\zposx{\theNoTableEntry r}sp},
          \n{urx}={0},
          \n{lly}={\zposy{\theNoTableEntry b}sp-\zposy{\theNoTableEntry r}sp},
          \n{ury}={\zposy{\theNoTableEntry t}sp-\zposy{\theNoTableEntry r}sp}
        in
        (\n{llx}, \n{lly}) -- (\n{urx}, \n{ury})
        (\n{llx}, \n{ury}) -- (\n{urx}, \n{lly})
      ;
    }%
  }%
}

\newcommand\T{\rule{0pt}{2.6ex}}        
\newcommand\B{\rule[-1.5ex]{0pt}{0pt}}

\definecolor{asparagus}{HTML}{A5DCAF}
\definecolor{cardinal}{HTML}{f88d87}
\definecolor{lg}{RGB}{255,255, 255}

\newcolumntype{b}{>{\columncolor{cardinal}}c}
\newcolumntype{a}{>{\columncolor{asparagus}}c}
\newcolumntype{g}{>{\columncolor{lg}}c}

\title{Cross-neutralising: Probing for joint encoding of linguistic information in multilingual models}

\author{Rochelle Choenni\\ ILLC, University of Amsterdam\\  \texttt{r.m.v.k.choenni@uva.nl} \And Ekaterina Shutova\\
  ILLC, University of Amsterdam \\
  \texttt{e.shutova@uva.nl} \\}

\date{}

\begin{document}
\maketitle
\begin{abstract}
  Multilingual text encoders are widely used to transfer NLP models across languages. The success of this transfer is, however, dependent on the model's ability to encode the patterns of cross-lingual similarity and variation. Yet, little is known as to how these models are able to do this. We propose a simple method to study how relationships between languages are encoded in two state-of-the-art multilingual models --- M-BERT and XLM-R. The results provide insight into their information sharing mechanisms and suggest that linguistic properties are encoded jointly across typologically-similar languages in these models.
\end{abstract}

\section{Introduction}

Early 
work in multilingual NLP focused on creating task-specific models, and can be divided into two main approaches: language transfer \citep{tackstrom2013target, tiedemann2014treebank} and multilingual joint learning \citep{ammar2016many, ammar2016massively, zhou2015subspace}. 
The former method enables the transfer of models or data from high to low-resource languages, hence porting information across languages, while the latter aims to leverage language interdependencies through joint learning from annotated examples in multiple languages.
Both methods relied on the fact that there are dependencies between processing different languages from a typological perspective. For instance, some syntactic properties are universal across languages (e.g. nouns take adjectives and determiners as dependents, but not adverbs), but others are influenced by the typological properties of each language (e.g. the order of these dependents with respect to the head) \citep{naseem2012selective}. We hypothesize that the pretrained general-purpose multilingual models (e.g. M-BERT \citep{devlin2019bert}) rely on these same concepts, and that some of the effectiveness of these models stems from the fact that they learn to efficiently encode and share information about linguistic properties across typologically-similar languages. In this paper, we examine cross-lingual interaction of linguistic information 
within M-BERT and XLM-R \citep{conneau2019unsupervised}, through the lens of typology. For instance, some shared properties of languages may be encoded jointly in the model, while others may be encoded separately in their individual subspaces. 

To investigate this, we develop a simple and yet novel method to probe for joint encoding of linguistic information, which we refer to as \textit{cross-neutralising}. Our work takes inspiration from \citet{choenni2020does}, who present a set of probing tasks to evaluate the extent to which multilingual models capture typological properties of languages, as defined in the World Atlas of Language Structures (WALS) \citep{wals}. We use the tasks introduced by \citet{choenni2020does}, but expand on their work by developing a method to probe for joint encoding of typological features. Previous research \cite{libovicky2020language,gonen2020s}  demonstrated that representations produced by M-BERT are projected to separate language-specific subspaces. Furthermore, they can be dissected into a language-neutral component, that captures the underlying meaning, and a language-specific component, that captures language identity. 
We exploit this property to test for information sharing between the language-specific subspaces
, and hypothesize that these subspaces jointly encode shared properties across typologically similar languages. 

To probe for joint encoding, we test to what extent
removing language-specific information negatively affects the probing classifier performance on the probing tasks in typologically-related languages. Our results show that by localizing information crucial for encoding the typological properties of one language, we are able to remove this same information from the representations of related languages (that share the same typological feature value). This indicates that the models jointly encode these typological properties across languages. 

\begin{table}
    \small
    \centering
    \begin{tabular}{|c|c|c|c|}
    \hline
    \textbf{Languages} \small{(ISO 639-1)} & \textbf{GN} & \textbf{NG} & \textbf{NDO}\\ 
     \hline
    da, hi, sv, mar & \notableentry & &\\
    \hline
    cs, mk, bg & & & \notableentry\\
    \hline
    pt, it, pl, es, fr & & \notableentry &\\
     \hline
    \end{tabular}
\caption{\footnotesize Probing task example of feature \texttt{86A}: \textit{Order of Genitive and Noun}. Labels are Genitive-Noun (GN), Noun-Genitive (NG) and No Dominant Order (NDO).}
\label{tab:taskexample}
\end{table}

\section{Related work}
Several works study language relationships within multilingual models. For instance, by reconstructing phylogenetic trees to analyze preserved relations 
(e.g. in terms of genetic and structural differences) \citep{ bjerva2019language, beinborn2019semantic}, or by probing for typological properties of languages \citep{qian2016investigating, csahin2019linspector, choenni2020does}. Our work comes closest to that of \citet{chi2020} who 
study shared grammatical relations in M-BERT. They use a structural probe \citep{hewitt2019} 
to enable zero-shot transfer across languages to successfully recover syntax. Their results suggest that the probe is able to pick up on features that are jointly encoded in M-BERT across languages. We expand on this work by linking these features to linguistic typology and demonstrating that individual lexical, morphological and syntactic properties of languages are jointly encoded across all languages that share the property. 

We draw inspiration from \citet{libovicky2020language} who show that M-BERT relies on a language-specific component that is similar across all representations in a language and can thus be approximated by its language centroid. They show that removing the respective centroid drastically decreases performance on language identification, while improving that 
on parallel sentence retrieval, indicating stronger language-neutrality. Hence, this method 
removes language-specific features from model representations 
(for M-BERT and XLM-R), while still encoding the underlying meaning. These results demonstrate the existence of the language-neutral component.
In subsequent work, \citet{gonen2020s} successfully decompose the representations into independent language-specific and language-neutral components through nullspace projections, thereby further supporting the existence of identifiable language components.

\begin{figure*}[t]
    \includegraphics[width=\linewidth, height=3.2cm]{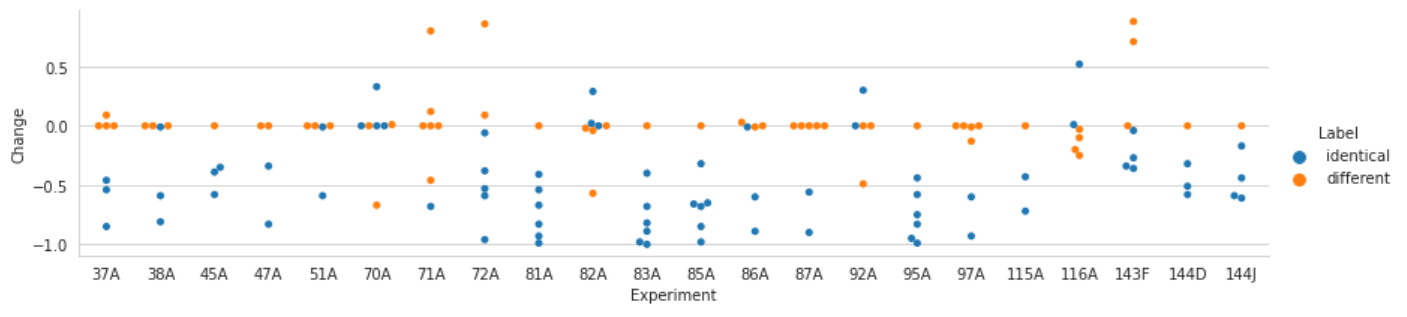}
    \vspace{-0.8cm}
    \caption{\small Change in performance for all test languages when cross-neutralising with Spanish. Languages are categorized by an identical (blue) or different (orange) feature value from Spanish for the respective task.}
    \label{fig:diff_plot}
    \vspace{-0.2cm}
\end{figure*}

\section{Multilingual models}
Both models are  12 layer bidirectional Transformers with 12 attention heads and a hidden state size of 768. We use the Multilingual Cased version of M-BERT that supports 104 languages, uses a 110K WordPiece vocabulary and is trained on Masked Language modelling (MLM) and Next Sentence Prediction (NSP). XLM-R is the multilingual variant of RoBerta \citep{liu2019roberta} that omits NSP and is trained with more data and compute power than M-BERT. It supports 100 languages and uses a Sentence Piece model (with a 250K vocabulary) for tokenization. To obtain fixed-length sentence representations, we perform mean-pooling over the hidden states of the top layer of the models.

\section{Methods}

\paragraph{Probing methods}
We use the 25 language-level probing tasks from \citet{choenni2020does} and adapt their paired language evaluation set-up using the same 7 typologically diverse language pairs: 1. (Russian, Ukrainian), 2. (Danish, Swedish), 3. (Czech, Polish), 4. (Portuguese, Spanish), 5. (Hindi, Marathi), 6. (Macedonian, Bulgarian), 7. (Italian, French). For each language, we retrieve 10K input sentences from the Tatoeba corpora\footnote{Tatoeba corpora available at: https://tatoeba.org} and annotate them with the WALS feature value for the corresponding language and probing task $\tau$. The 25 features we probe for span a wide range of linguistic properties pertaining to lexical, morphological and syntactic structure, classified under the following codes and categories\footnote{For full feature names see Appendix \ref{app:details} Table \ref{app:tablenames}, for descriptions see: https://wals.info/}: {[37A, 38A, 45A, 47A, 51A]} (Nominal), {[70A, 71A, 72A, 79A, 79B]}  (Verbal), {[81A, 82A, 83A, 85A, 86A, 87A, 92A, 93A, 95A, 97A, 143F, 144D, 144J]} (Word order) and {[115A, 116A]} (Simple clauses). See Table \ref{tab:taskexample} for an example of a probing task.

For each 
pair, we use the first language for training and the second for testing. 
Note that not all features have annotations for all languages, in which case we omit the language from the task. Thus, given that a feature covers $n$ language-pairs, we train the probing classifier on $n \times 10K$ sentences from all of the training languages. 
Following \citet{choenni2020does}, we use a one-layer MLP with 100 hidden units, ReLU activation, and an output layer that uses the softmax function to predict the feature values. The parameters of the sentence encoder are frozen during training such that all learning can be ascribed to the probing classifier $P_{\tau}$. We then use $P_{\tau}$ to predict the feature values for the $n\times 10K$ test sentences. For more details of the training regime, see Appendix \ref{app:details}.

\paragraph{Restructuring the vector space}
 Following \citet{libovicky2020language}, we approximate the language centroid for each language in our test set $x \in \mathrm{L}$, by obtaining a mean language vector $\mathbf{\bar{u}_x} \in \mathbb{R}^{m}$ from a set of $N$ sentence representations $u_1, .. , u_N \in \mathbb{R}^{m}$ from that language (10K in our case). 
 The idea is that by localizing language-specific information through averaging representations, core linguistic properties remain prominent in the centroid. Simultaneously, infrequent phenomena that vary depending on sentence meaning are averaged out. We then obtain a set of language-neutral representations $v_i \in \mathbb{R}^{m}$ for a language $x$ by subtracting the corresponding language centroid from the 
 model representation $h_i$ for a sentence $i$: $v_{i} = h_{i} - \mathbf{\bar{u}_x}$.
\noindent This means that we remove language-specific information by re-structuring the vector space such that the average of the representations for each language is centered at the origin of the vector space. From now on we refer to this method as \textit{self-neutralising}.

\paragraph{Testing for information-sharing}
 
To investigate how typological properties are shared, i.e. whether they are jointly encoded across languages in a localizable manner or rather in independent ways for each language, we adapt this method to a cross-neutralising scenario. Specifically, we approximate typological information from one language ($x$) by computing $\mathbf{\bar{u}_x}$, and subtract $\mathbf{\bar{u}_x}$ from the representations of all languages in $\mathrm{L} \setminus \{x\} $. We then test the trained probing classifier on the neutralised representations of each language. 
If the encoders were to represent languages and their properties in independent ways, we expect the probing performance to deteriorate only for language $x$. In case of joint encoding of typological properties, however, we expect to see that 
performance (1) also deteriorates for other languages that share the same typological feature value with $x$ and (2) remains intact for languages that do not share the same feature value with $x$. 
We refer to this method as \textit{cross-neutralising}. 

\section{Experiments and results}

\begin{table*}[t]
\vspace{-0.5cm}
\setlength{\tabcolsep}{5pt}
\scriptsize
\centering
\begin{tabular}{c|ba|ba|ba|ba|ba|ba|ba}
\hline
\T
\multirow{2}{*}{\diagbox{$\tau$}{$x$}}& \multicolumn{2}{c|}{Ukrainian} &\multicolumn{2}{c|}{Swedish} &\multicolumn{2}{c|}{Polish} &\multicolumn{2}{c|}{Spanish} &\multicolumn{2}{c|}{Marathi} &\multicolumn{2}{c|}{Bulgarian} &\multicolumn{2}{c}{French} \\
&  same & diff. & same & diff. &same & diff. &same & diff. &same & diff. &same & diff. &same & diff.  \\
\hline
      \T
       37A &  -0.45&-0.22 &  -0.76&-0.01 &  -0.62&0.03 &   -0.62&0.02 &   -0.74&0.04 &   -0.46&0.02 &   -0.15&0.01 \\
       38A &    -0.8&0.0 &     \textcolor{gray}{0.11}&\textcolor{gray}{0.0} &   -0.67&0.0 &    -0.47&0.0 &   -0.82&0.05 &  \multicolumn{2}{g|}{--}&    -0.34&0.0 \\
       45A & \multicolumn{2}{g|}{--}&  -0.4&0.0 &           \multicolumn{2}{g|}{--}&    -0.44&0.0 &    -0.61&0.0 &            \multicolumn{2}{g|}{--}&    -0.21&0.0 \\
       47A & \multicolumn{2}{g|}{--}&   -0.32&0.01 &           \multicolumn{2}{g|}{--}&    -0.58&0.0 &   -0.82&0.01 &            \multicolumn{2}{g|}{--}&    -0.18&0.0 \\
       51A &   -0.22&0.14 &   \textcolor{gray}{0.18}&\textcolor{gray}{-0.23} &  -0.26&0.17 &     -0.3&0.0 &   -0.75&0.44 &  \multicolumn{2}{g|}{--}& \multicolumn{2}{g}{--}\\
       70A &   -0.58&0.33 &    -0.38&0.0 &  -0.26&0.12 &  \textcolor{gray}{ 0.08}&\textcolor{gray}{-0.22} &  \textcolor{gray}{-0.1}&\textcolor{gray}{-0.54} &   -0.61&0.21 &   -0.51&0.05 \\
       71A &   -0.15&0.04 &   -0.55&0.05 &  -0.45&0.05 &   -0.68&0.08 &   \textcolor{gray}{0.08}&\textcolor{gray}{-0.06} & -0.28&-0.12 &  \textcolor{gray}{-0.13}&\textcolor{gray}{-0.11} \\
       72A &   -0.07&0.05 &   -0.36&0.39 &  -0.35&0.39 &    -0.5&0.48 &   \textcolor{gray}{0.24}&\textcolor{gray}{-0.18} &    -0.23&0.0 &   -0.81&0.54 \\
 79A &  -0.2&0.05 &  \multicolumn{2}{g|}{--}&  -0.68&0.07 &            \multicolumn{2}{g|}{--}&  \multicolumn{2}{g|}{--}&   -0.43&0.01 & \multicolumn{2}{g}{--}\\
       79B &    \textcolor{gray}{0.14}&\textcolor{gray}{0.07} & \multicolumn{2}{g|}{--}&  -0.35&0.06 &            \multicolumn{2}{g|}{--}&  \multicolumn{2}{g|}{--}&   -0.45&0.13 & \multicolumn{2}{g}{--}\\
       81A &   -0.1&0.0 &    -0.62&0.0 &   -0.28&0.0 &    -0.73&0.0 &    -0.57&0.0 &   -0.46&0.0 &    -0.38&0.0 \\
       82A &    -0.4&0.35 &   -0.39&0.32 &  \textcolor{gray}{0.32}&\textcolor{gray}{-0.36} &    \textcolor{gray}{0.1}&\textcolor{gray}{-0.16} &   -0.53&0.42 &   \textcolor{gray}{0.71}&\textcolor{gray}{-0.82} &   -0.06&0.03 \\
       83A &    -0.08&0.0 &     -0.6&0.0 &   -0.27&0.0 &     -0.8&0.0 &     -0.6&0.0 &    -0.41&0.0 &    -0.47&0.0 \\
       85A &    -0.07&0.0 &    -0.41&0.0 &   -0.27&0.0 &    -0.69&0.0 &    -0.64&0.0 &     -0.4&0.0 &    -0.38&0.0 \\
       86A & \multicolumn{2}{g|}{--}&    -0.34&0.0 &  \textcolor{gray}{0.09}&\textcolor{gray}{-0.11} &    -0.5&0.01 &    -0.84&0.0 &    -0.6&0.06 &   -0.42&0.01 \\
       87A &   -0.29&0.02 &   -0.32&0.02 &  -0.16&0.02 &    -0.73&0.0 &    -0.8&0.02 &   -0.67&0.02 &    -0.34&0.0 \\
       92A &  \textcolor{gray}{ 0.26}&\textcolor{gray}{-0.05} &   -0.35&0.15 &  \textcolor{gray}{0.28}&\textcolor{gray}{-0.35} &  \textcolor{gray}{ 0.15}&\textcolor{gray}{-0.16} &         \multicolumn{2}{g|}{--}&  \multicolumn{2}{g|}{--}&    \textcolor{gray}{-0.06}&\textcolor{gray}{0.1} \\
       93A & \multicolumn{2}{g|}{--}&    -0.29&0.0 &  \textcolor{gray}{0.19}&\textcolor{gray}{-0.21} &            \multicolumn{2}{g|}{--}&  -0.53 & 0.13&            \multicolumn{2}{g|}{--}&  \multicolumn{2}{g}{--}\\
       95A &    -0.12&0.0 &    -0.65&0.0 &   -0.28&0.0 &    -0.76&0.0 &    -0.53&0.0 &    -0.47&0.0 &    -0.43&0.0 \\
       97A &    -0.24&0.0 &     -0.7&0.0 &   -0.48&0.0 &  -0.76&-0.03 &  -0.72&-0.03 &    -0.87&0.0 &  -0.42&-0.02 \\
      115A & \multicolumn{2}{g|}{--}&  \multicolumn{2}{g|}{--}&           \multicolumn{2}{g|}{--}&  -0.57&0.0 &  -0.55&0.0 &            \multicolumn{2}{g|}{--}&  -0.28&0.0 \\
      116A &    -0.47&0.4 &   -0.22&0.19 &  \textcolor{gray}{0.11}&\textcolor{gray}{-0.02} &   \textcolor{gray}{0.26}&\textcolor{gray}{-0.14} &   -0.27&0.28 &   -0.36&0.34 & \multicolumn{2}{g}{--}\\
      143F &   -0.75&0.66 &    -0.21&0.0 &  -0.23&0.52 &   -0.25&0.53 &   \textcolor{gray}{0.24}&\textcolor{gray}{-0.03} &   -0.25&0.53 &   \textcolor{gray}{0.16}&\textcolor{gray}{-0.02} \\
      144D &    -0.62&0.0 &  -0.47&0.0 &  \multicolumn{2}{g|}{--}&    -0.47&0.0 &   \multicolumn{2}{g|}{--}&   -0.34&0.0 & \multicolumn{2}{g}{--}\\
      144J & -0.74&0.0 &   -0.54&0.0 &   -0.32&0.0 &    -0.45&0.0 &            \multicolumn{2}{g|}{--}&    -0.28&0.0 & \multicolumn{2}{g}{--}
\end{tabular}
\vspace{-0.1cm}
\caption{\small The average change in performance per task $\tau$ and cross-neutralizing language $x$ for M-BERT, categorized by languages that have the same (same) or different (diff) feature value from language $x$. Cases for which the probing task performance on the language before neutralising was insufficient ($< 75\%$ accuracy) are denoted in gray (it is unclear what information these centroids capture, hence we can not reasonably expect the same trend to emerge). Note, the blank spaces indicate the cases in which $x$ was omitted from the task due to a lack of coverage in WALS.}
\label{tab:subset_diff-neut-bert}
\vspace{-0.2cm}
\end{table*}


\subsection{Self-neutralising}\label{sec:lc}

First, we test 
whether our approximated language centroids $\mathbf{\bar{u}_x}$ successfully capture the typological properties of the language. We do this by testing whether self-neutralising results in a substantial loss of information about the typological properties of the languages in our test set. We evaluate the change in probing performance before and after applying this method, and observe that self-neutralising decreases performance to chance accuracy for each language (Appendix \ref{app:before-after-self}, Tab. \ref{tab:self}). Thus, the method successfully removes crucial typological information from the encodings. Moreover, the language identity, approximated by the language centroid, is crucial for the encoding of typological properties, suggesting that typological information is largely encoded in the relative positioning of the language-specific subspaces of our models. 


\subsection{Cross-neutralising}\label{sec:cross-neut}

Now that we know that computing $\mathbf{\bar{u}_x}$ is a viable method to localise the typological properties of a language $x$, we apply our cross-neutralising method. From the results we see that depending on the language we cross-neutralise with (i.e. language $x$ from which we compute $\mathbf{\bar{u}_x}$): 1. performance on a different set of languages is affected, and 2. this set of languages varies per task. Upon further inspection, we observe that the affected languages tend to share the same feature value as $x$ for the respective task. Figure \ref{fig:diff_plot} shows the change in performance on all test languages when cross-neutralised with Spanish (see Appendix \ref{app:cross-neut-change-test-langs} for cross-neutralisation with other languages). We categorize these languages based on whether their feature value is the same (blue) or different (orange) from the feature value of Spanish in the respective task. We indeed see that the performance on the set of languages that have the same feature value tend to deteriorate, while the performance on languages with a different feature value remains mostly constant. 

Moreover, 
when the probe predicts the incorrect feature value for language $x$, we find that the languages that share this value are affected instead (regardless of typological relationship). For instance, for task \texttt{116A} : `\textit{Polar Questions}' the label `\textit{Question particle}' is always incorrectly predicted for the Spanish representations (even before neutralising). Consequently, when cross-neutralising with Spanish, the performance for languages that share this feature value deteriorates (note that in Fig. \ref{fig:diff_plot} the orange dots drop in this case). 
This indicates that the model encodes the feature value `\textit{Question particle}' for Spanish. Thus, when we compute $\mathbf{\bar{u}_x}$, we capture information about this feature value instead of the correct one `\textit{Interrogative word order}'. 

Table \ref{tab:subset_diff-neut-bert} shows the average change in performance for M-BERT, categorized by feature value, for each language with which we neutralise (see Appendix \ref{app:full_tables_avg}, Table \ref{tab:diff-neut-xlm} for XLM-R results). 
The table shows that there is a clear overall pattern where the performance in languages with the same feature value suffers, while that in languages with a different feature value remains intact. 
These results hold true for all languages we cross-neutralise with and for both encoders. 
In some cases, however, we notice that cross-neutralising on average increases performance in languages with a different feature value (e.g. $x$ = Ukrainian for task \texttt{70A}). We speculate that removing information about the feature value of $x$ reduces noise in the representations allowing the probe to pick up on the right signal. 

Thus, we find that language centroids capture specific feature values in a localizable and systematically similar way across different languages, indicating that typological properties are jointly encoded across languages. We re-produced all our experiments using sentence representations from the other layers of the models and obtained similar results in all layers (see Appendix \ref{app:cross-all-layers}, Fig. \ref{fig:cross_across_layers}). 

\section{Conclusion}

We have shown that typological feature values are encoded jointly across languages and are localizable in their respective language centroids. In the future, we will correlate the model’s ability to encode typological features with its performance in downstream tasks by progressively deteriorating the amount of typological information encoded. 
Moreover, our method enables us to carefully 
select which 
languages we want to neutralise w.r.t. 
 certain typological properties. This could inspire work on encouraging selective generalization in large-scale models based on typological knowledge, as opposed to enforcing complete language-agnosticism. Lastly, our method is easily applicable to probing for joint encoding in other scenarios, e.g. linguistic and visual information in multimodal models. 

\onecolumn

\appendix

\section{Reproducibility details}\label{app:details}

\begin{table}[H]
\begin{center}
     \begin{tabular}{|l|l|l|}
        \hline 
        \textbf{Code}  & \textbf{Category} & \textbf{Feature name} \\
        \hline
        37A& Nominal Category & Definite articles\\
        38A*& Nominal Category & Indefinite articles\\
        45A$^{\dagger}$ & Nominal Category & Politeness distinctions in pronouns\\
        47A$^{\dagger}$  & Nominal Category & Intensifiers and reflexive pronouns \\
        \B 51A $^{\ddagger}$ & Nominal Category &Position of case affixes  \\
        \hline
        70A &Verbal Category& The morphological imperative \T\\
        71A &Verbal Category& The prohibitive\\
        72A &Verbal Category& Imperative-hortative systems\\ 
        79A$^{\mathsection}$ &Verbal Category& Suppletion according to tense and aspect \\
        \B 79B$^{\mathsection}$  &Verbal Category& Suppletion in imperatives and hortatives \\
        \hline
        \T 81A & Word Order & Order of Subject, Object and Verb (SOV)\\
        82A & Word Order & Order of Subject and Verb (SV) \\
        83A & Word Order & Order of Object and Verb (OV)\\
        85A & Word Order &Order of adposition and noun phrase\\
        86A$^{\dagger}$ & Word Order & Order of genitive and noun \\
        87A & Word Order &Order of adjective and noun\\
        92A$^{\mid}$ & Word Order &Position of polar question particles \\
        93A$^{\mathparagraph}$  & Word Order &Position of interrogative phrases in content questions\\
        95A & Word Order & Relationship between OV and adposition and noun phrase order\\
        \B 97A & Word Order &Relationship between OV and adjective and noun order \\
        \hline 
        \T 115A$^{\#}$ & Simple Clauses & Negative indefinite pronouns and predicate negation\\
        116A$^{\lozenge}$ & Simple Clauses & Polar questions \B \\
        \hline 
        143F & Word Order &Postverbal negative morphemes \T \\
        144D$^{\downarrow}$ & Word Order & Position of negative morphemes\\ 
        \B 144J$^{\delta}$  & Word Order & Order of Subject, Verb, Negative word, and Object (SVNegO)\\
        \bottomrule
      \end{tabular}
      \caption[Specifications of the 25 WALS features used for probing]{The 25 WALS features used for probing along with their correpsonding WALS codes and categories. The multilingual sentence representations for each of these features are probed for in separate tasks. Unless indicated otherwise, all language pairs were covered. Excluded pairs: *:(1), $^{\dagger}$:(1, 3 and 6), 
$^{\ddagger}$:(6 and 7), $^{\mathsection}$:( 2, 4, 5 and 7), $^{\mid}$:(5 and 6), $^{\mathparagraph}$:(1, 4, 6, 7), $^{\#}$:(1-3 and 6), $^{\lozenge}$:(7), $^{\downarrow}$:(3, 5 and 7), $^{\delta}$:(5 and 7)} 
     \label{app:tablenames}
    \end{center}
\end{table}

\paragraph{Probing tasks} 
Table \ref{app:tablenames} contains the WALS codes, categories and feature names of the 25 WALS features used for our probing tasks. For detailed descriptions of these features the reader is referred to the original documentations at: \url{https://wals.info/}. Also, please note the indication of excluded language pairs per task.

\paragraph{Training regime} No fine-tuning is performed on the hyperparameters of the probing classifier to keep results across tasks comparable. For each task we trained for 20 epochs, with early stopping (patience=5), using the Adam optimizer \citep{kingma2014adam}.  We set the batch size to 32 and use a dropout rate of 0.5. Note that not all tasks are binary classification problems, hence we use one-hot label encodings and return the class with the highest probability at test time.

\section{Results before and after self-neutralising}\label{app:before-after-self}
\begin{table}[H]
\begin{center}
     \begin{tabular}{c|c|c|c|c}
     & \multicolumn{2}{c|}{M-BERT} & \multicolumn{2}{c}{XLM-R} \\
     \hline
     $\tau$ & before & after & before & after \\
        \hline 
        37A & 0.96 $\pm$ 0.04 & 0.4 $\pm$ 0.07& 1.0 $\pm$ 0.01& 0.41 $\pm$ 0.06 \\
        38A & 0.83 $\pm$ 0.37 & 0.37 $\pm$ 0.03 & 0.83$\pm$ 0.37 & 0.39 $\pm$ 0.03\\
        45A & 1.0 $\pm$ 0.0 & 0.58 $\pm$ 0.11& 1.0 $\pm$ 0.0 &0.54 $\pm$ 0.04 \\
        47A & 1.0 $\pm$ 0.001 & 0.51 $\pm$  0.14& 1.0 $\pm$ 0.0 & 0.5$\pm$ 0.04 \\
        51A & 0.8 $\pm$ 0.4 & 0.53 $\pm$ 0.11 & 0.8 $\pm$ 0.4& 0.5 $\pm$ 0.04 \\
        70A & 0.71 $\pm$ 0.45 & 0.34 $\pm$ 0.09 & 0.71 $\pm$ 0.45 & 0.38 $\pm$ 0.11 \\
        71A & 0.69 $\pm$ 0.43 & 0.36 $\pm$ 0.1 & 0.76 $\pm$ 0.37  & 0.35 $\pm$ 0.08 \\
        72A & 0.84 $\pm$ 0.35 & 0.56 $\pm$ 0.09 & 0.84 $\pm$ 0.35& 0.58 $\pm$ 0.09\\
        79A & 0.95 $\pm$ 0.05 & 0.54 $\pm$ 0.04 &0.98$\pm$ 0.03  & 0.55 $\pm$ 0.03 \\
        79B & 0.63 $\pm$ 0.45 & 0.33 $\pm$  0.07 & 0.65 $\pm$ 0.46 &0.38 $\pm$ 0.1 \\
        81A &  1.0 $\pm$ 0.0 & 0.57 $\pm$ 0.06 & 1.0$\pm$ 0.01 & 0.53 $\pm$ 0.04 \\
        82A & 0.53 $\pm$ 0.47 & 0.53 $\pm$ 0.02 & 0.48 $\pm$ 0.45 & 0.5 $\pm$ 0.08 \\
        83A & 1.0 $\pm$ 0.0 & 0.58 $\pm$ 0.07 & 1.0 $\pm$ 0.0 & 0.5 $\pm$ 0.02 \\
        85A & 1.0 $\pm$ 0.0 & 0.63 $\pm$ 0.11& 1.0 $\pm$ 0.01 & 0.52 $\pm$ 0.02 \\
        86A & 0.82 $\pm$ 0.37 & 0.35 $\pm$  0.04& 0.83 $\pm$  0.36& 0.36 $\pm$ 0.01 \\
        87A & 1.0 $\pm$ 0.0 & 0.54 $\pm$ 0.06& 1.0 $\pm$ 0.0&0.52 $\pm$ 0.03 \\
        92A & 0.24 $\pm$ 0.38 & 0.37 $\pm$ 0.02& 0.28 $\pm$ 0.39 & 0.36 $\pm$ 0.05 \\
        93A & 0.65 $\pm$ 0.46 & 0.42 $\pm$ 0.03 & 0.67 $\pm$ 0.47 &0.48 $\pm$ 0.03 \\
        95A & 1.0 $\pm$ 0.0 & 0.54 $\pm$  0.03& 1.0 $\pm$ 0.0 & 0.5 $\pm$ 0.01 \\
        97A & 1.0 $\pm$ 0.01 & 0.39 $\pm$ 0.05 & 1.0 $\pm$ 0.0 & 0.4 $\pm$ 0.08 \\
        115A & 1.0 $\pm$ 0.0 & 0.53 $\pm$ 0.05& 1.0 $\pm$ 0.0 & 0.52 $\pm$ 0.03 \\
        116A & 0.67 $\pm$ 0.47 & 0.5 $\pm$ 0.03 & 0.68 $\pm$ 0.45&0.5 $\pm$ 0.03 \\
        143F & 0.71 $\pm$ 0.45 & 0.52 $\pm$ 0.14& 0.71 $\pm$ 0.45 & 0.52 $\pm$ 0.07 \\
        144D & 1.0 $\pm$ 0.0 & 0.5 $\pm$ 0.02 & 1.0 $\pm$ 0.0 & 0.52 $\pm$ 0.03 \\
        144J & 1.0 $\pm$ 0.0 & 0.54 $\pm$ 0.04 & 1.0 $\pm$ 0.0 & 0.54 $\pm$ 0.06 \\
    \end{tabular}
     \caption{The table shows the mean and standard deviation of the performance in (\%) accuracy computed across languages in the test set. We report the results obtained before any neutralisation and after self-neutralising each language. }
          \label{tab:self}
\end{center}
\end{table}
\newpage

\section{Cross-neutralising results for M-BERT}\label{app:cross-neut-change-test-langs}

\begin{figure}[H]
    \centering
    \includegraphics[width=\linewidth, height=3.5cm]{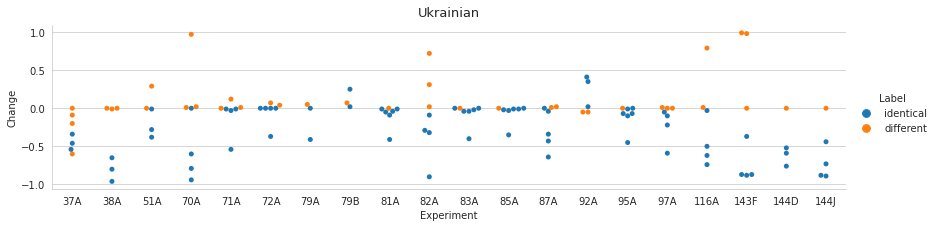}
       \includegraphics[width=\linewidth, height=3.5cm]{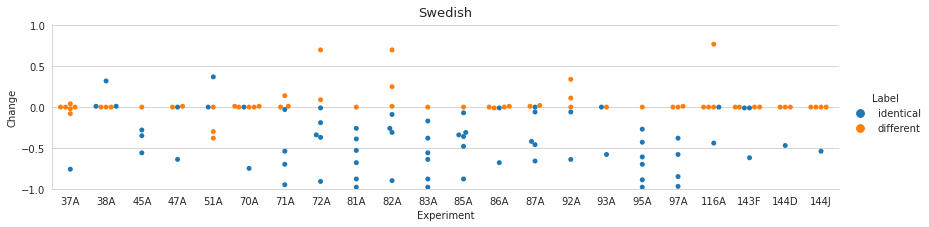}
        \includegraphics[width=\linewidth, height=3.5cm]{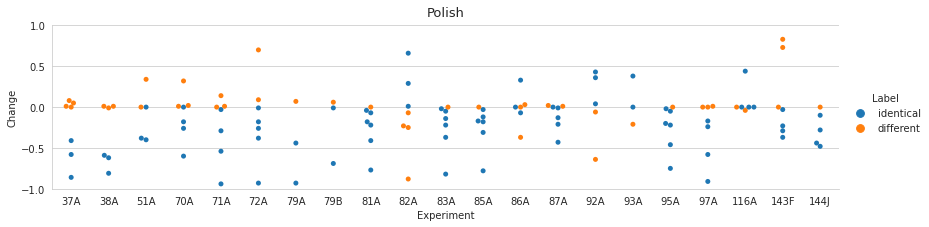}
        \includegraphics[width=\linewidth, height=3.5cm]{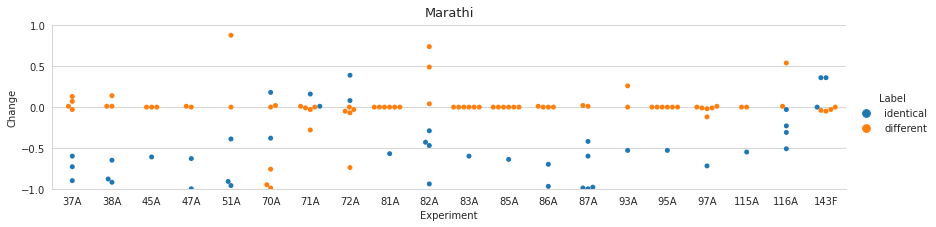}
        \includegraphics[width=\linewidth, height=3.5cm]{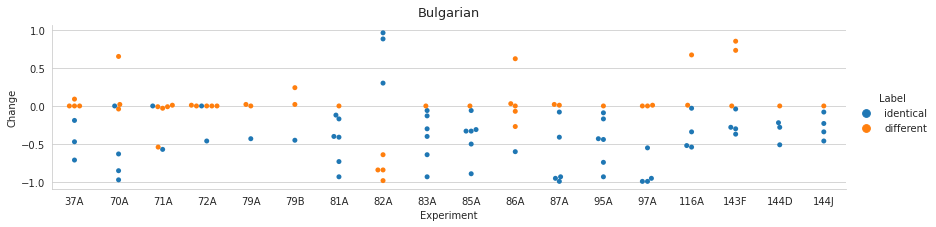}
        \includegraphics[width=\linewidth, height=3.5cm]{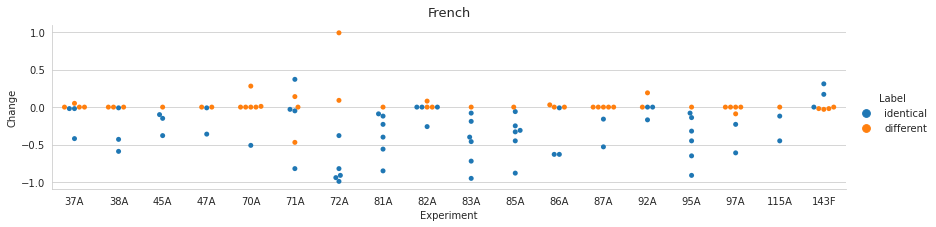}
    \caption{Change in performance after cross-neutralising with the other test languages for M-BERT. The performance change for all 25 probing tasks is shown per language used for cross-neutralising. }
    \label{fig:neut_diff_heatmaps}
\end{figure}

\section{Averaged performance change over languages for XLM-R}\label{app:full_tables_avg}

\begin{table}[H]
\scriptsize
\setlength{\tabcolsep}{5pt}
\centering
\begin{tabular}{c|ba|ba|ba|ba|ba|ba|ba}
\hline
\T
\multirow{2}{*}{\diagbox{$\tau$}{$x$}}& \multicolumn{2}{c|}{Ukrainian} &\multicolumn{2}{c|}{Swedish} &\multicolumn{2}{c|}{Polish} &\multicolumn{2}{c|}{Spanish} &\multicolumn{2}{c|}{Marathi} &\multicolumn{2}{c|}{Bulgarian} &\multicolumn{2}{c}{French} \\
&  same & diff. & same & diff. &same & diff. &same & diff. &same & diff. &same & diff. &same & diff.  \\
\hline
\T
   37A &    -0.26&0.0 &  -0.74&-0.02 &  -0.77&0.01 &   -0.81&0.0 &  -0.74&0.01 &   -0.54&0.0 &    -0.22&0.0 \\
       38A &   -0.76&0.02 &   \textcolor{gray}{ 0.13}&\textcolor{gray}{-0.0} &   -0.5&0.01 &   -0.52&0.0 &   -0.88&0.0 & \multicolumn{2}{g|}{--}&      -0.2&0.0 \\
       45A & \multicolumn{2}{g|}{--}&    -0.72&0.0 &            \multicolumn{2}{g|}{--}&   -0.34&0.0 &   -0.52&0.0 &            \multicolumn{2}{g|}{--}&    -0.28&0.0 \\
       47A &     \multicolumn{2}{g|}{--}&    -0.48&0.0 &            \multicolumn{2}{g|}{--}&   -0.72&0.0 &   -0.64&0.0 &            \multicolumn{2}{g|}{--}&    -0.24&0.0 \\
       51A &   -0.81&0.51 &  \textcolor{gray}{ 0.24}&\textcolor{gray}{-0.14} &  -0.17&0.31 &   -0.27&0.0 &   -0.53&0.5 & \multicolumn{2}{g|}{--}&     \multicolumn{2}{g}{--}\\
       70A &   -0.63&0.02 &  -0.36&-0.01 &  -0.16&0.02 &  \textcolor{gray}{0.11}&\textcolor{gray}{-0.31} &  \textcolor{gray}{ 0.1}&\textcolor{gray}{-0.57} &   -0.6&0.03 &   -0.62&0.02 \\
       71A &  -0.66&-0.08 &    -0.68&0.0 &  -0.72&0.01 &  -0.68&0.11 &    \textcolor{gray}{0.08}&\textcolor{gray}{0.0} &  -0.27&0.04 &  \textcolor{gray}{-0.18}&\textcolor{gray}{-0.19} \\
       72A &   -0.09&0.03 &   -0.32&0.04 &  -0.77&0.39 &  -0.33&0.04 &  \textcolor{gray}{0.24}&\textcolor{gray}{-0.61} &    -0.2&0.0 &   -0.69&0.28 \\
       79A &   -0.24&0.05 &  \multicolumn{2}{g|}{--}&  -0.72&0.07 &   \multicolumn{2}{g|}{--}&    \multicolumn{2}{g|}{--}&  -0.35&0.01 & \multicolumn{2}{g}{--}\\
       79B &   \textcolor{gray}{0.29}&\textcolor{gray}{-0.03} &  \multicolumn{2}{g|}{--}&  -0.22&0.03 &            \multicolumn{2}{g|}{--}&    \multicolumn{2}{g|}{--}&  -0.44&0.13 & \multicolumn{2}{g}{--}\\
       81A &    -0.09&0.0 &    -0.81&0.0 &   -0.42&0.0 &   -0.51&0.0 &    -0.5&0.0 &   -0.57&0.0 &    -0.52&0.0 \\
       82A &    -0.52&0.7 &   -0.72&0.73 &  \textcolor{gray}{0.29}&\textcolor{gray}{-0.25} &  \textcolor{gray}{0.25}&\textcolor{gray}{-0.24} &  -0.48&0.63 &   \textcolor{gray}{0.8}&\textcolor{gray}{-0.84} &  \textcolor{gray}{ -0.08}&\textcolor{gray}{0.02} \\
       83A &    -0.14&0.0 &    -0.81&0.0 &   -0.41&0.0 &   -0.52&0.0 &   -0.48&0.0 &   -0.52&0.0 &    -0.55&0.0 \\
       85A &    -0.19&0.0 &    -0.84&0.0 &   -0.32&0.0 &   -0.49&0.0 &   -0.49&0.0 &   -0.47&0.0 &    -0.55&0.0 \\
       86A & \multicolumn{2}{g|}{--}&    -0.58&0.1 &  \textcolor{gray}{0.09}&\textcolor{gray}{-0.02} &   -0.56&0.0 &  -0.8&-0.02 &  -0.62&0.06 &   -0.3&-0.01 \\
       87A &   -0.88&0.01 &   -0.28&0.01 &  -0.19&0.01 &   -0.75&0.0 &  -0.58&0.01 &  -0.44&0.01 &    -0.28&0.0 \\
       92A &    \textcolor{gray}{0.7}&\textcolor{gray}{-0.26} &   -0.38&0.26 &   \textcolor{gray}{0.4}&\textcolor{gray}{-0.12} & \textcolor{gray}{  0.13}&\textcolor{gray}{-0.0} &            \multicolumn{2}{g|}{--}&    \multicolumn{2}{g|}{--}&   \textcolor{gray}{0.1}&\textcolor{gray}{-0.08} \\
       93A &  \multicolumn{2}{g|}{--}&    -0.26&0.0 &    \textcolor{gray}{0.22}&\textcolor{gray}{0.0} &            \multicolumn{2}{g|}{--}&   -0.49& 0.5&            \multicolumn{2}{g|}{--}&   \multicolumn{2}{g}{--}\\
       95A &   -0.13&0.01 &   -0.83&0.01 &  -0.43&0.01 &  -0.52&0.01 &   -0.48&0.0 &  -0.56&0.01 &   -0.52&0.01 \\
       97A &   -0.89&0.03 &   -0.69&0.03 &  -0.22&0.03 &   -0.7&0.01 &  -0.72&-0.0 &  -0.64&0.03 &  -0.26&-0.01 \\
      115A &             \multicolumn{2}{g|}{--}&             \multicolumn{2}{g|}{--}&            \multicolumn{2}{g|}{--}&    -0.6&0.0 &   -0.51&0.0 &            \multicolumn{2}{g|}{--}&    -0.32&0.0 \\
      116A &   -0.14&0.02 &   -0.26&0.23 &  \textcolor{gray}{0.11}&\textcolor{gray}{-0.02} &  \textcolor{gray}{0.22}&\textcolor{gray}{-0.32} &  -0.56&0.49 &  -0.45&0.45 &             \multicolumn{2}{g}{--}\\
      143F &    -0.13&0.2 &    -0.19&0.0 &   -0.3&0.29 &  -0.67&0.34 &  \textcolor{gray}{0.48}&\textcolor{gray}{-0.97} &  -0.74&0.34 &   \textcolor{gray}{0.14}&\textcolor{gray}{-0.09} \\
      144D &    -0.16&0.0 &    -0.52&0.0 &            \multicolumn{2}{g|}{--}&   -0.53&0.0 &            \multicolumn{2}{g|}{--}&   -0.75&0.0 &             \multicolumn{2}{g}{--}\\
      144J &    -0.14&0.0 &    -0.58&0.0 &   -0.36&0.0 &   -0.55&0.0 &            \multicolumn{2}{g|}{--}&   -0.81&0.0 &             \multicolumn{2}{g}{--}\\
\end{tabular}
\caption{\small The average change in performance per task $\tau$ and cross-neutralizing language $x$ for XLM-R categorized by languages that have the same and those that have a different feature value from language $x$. Cases for which the probe performance on the language before neutralising was insufficient ($< 75\%$ accuracy) are denoted in gray (it is unclear what information these centroids capture, hence we can not reasonably expect the same trend to emerge). Note, the blank spaces indicate the cases in which $x$ was omitted from the task due to a lack of coverage in WALS. }
\label{tab:diff-neut-xlm}
\end{table}

\section{Cross-neutralising results for M-BERT across layers}\label{app:cross-all-layers}

\begin{figure}[H]
    \centering
    \includegraphics[width=\linewidth, height=10cm]{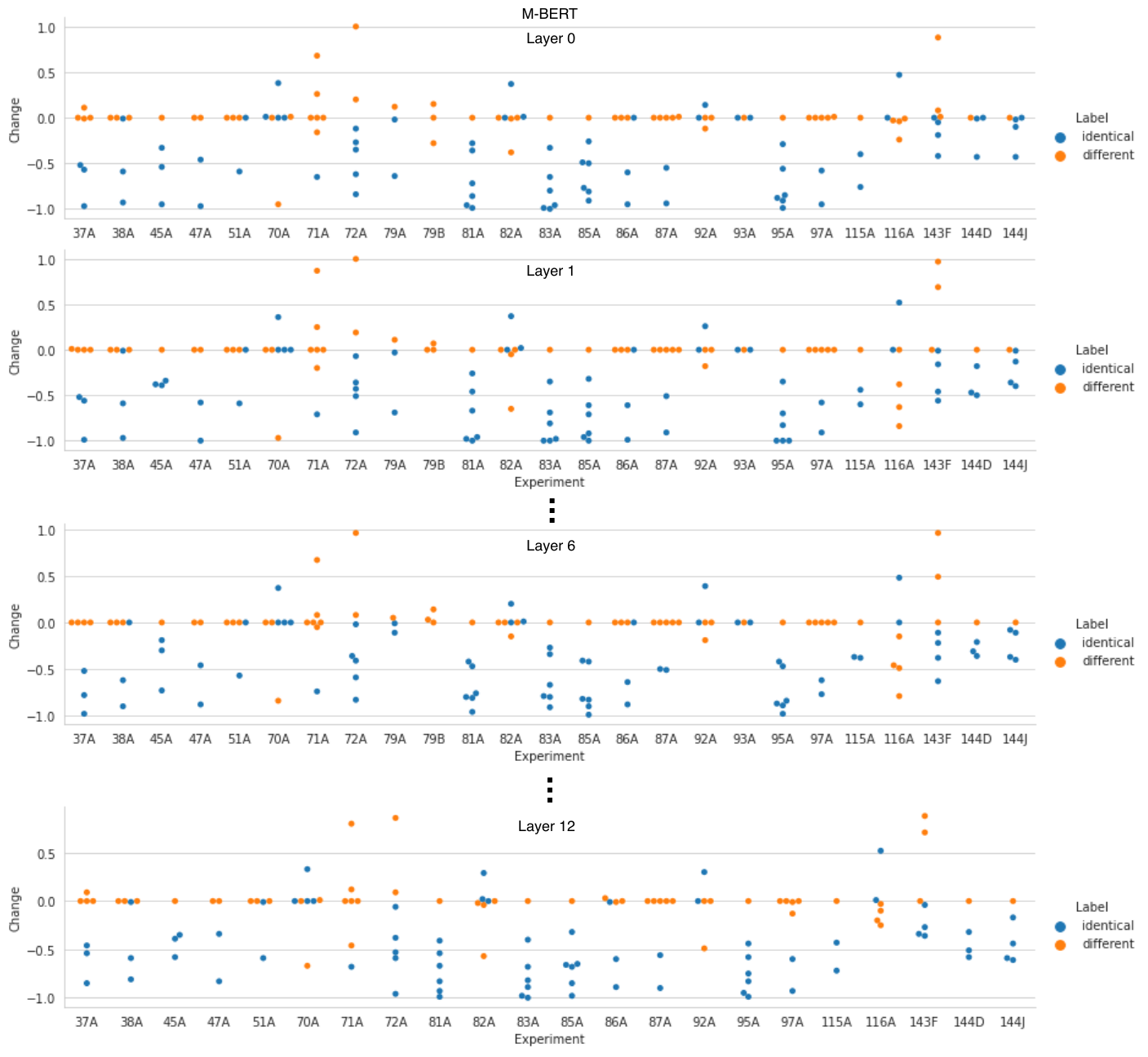}
    \caption{\small The change in performance for all test languages when cross-neutralising M-BERT representations with a language-centroid computed from the Spanish sentences. Languages are categorized by whether they had the same or a different feature value from that of Spanish for the respective tasks.}
    \label{fig:cross_across_layers}
\end{figure}
\newpage

\end{document}